\documentclass[11pt,a4paper]{article}
\usepackage[hyperref]{acl2020}
\usepackage{times}
\usepackage{latexsym}

\usepackage{microtype}
\usepackage{ amssymb }
\usepackage{amsmath}
\usepackage{booktabs}
\usepackage{multirow}
\usepackage{algorithm}
\usepackage{caption}
\usepackage{algorithmic}

\usepackage{fixltx2e}

\usepackage{bbm}
\usepackage{bm}

\makeatletter
\newcommand{\thickhline}{%
    \noalign {\ifnum 0=`}\fi \hrule height 2pt
    \futurelet \reserved@a \@xhline
}
\makeatother
\usepackage{tabularx}
\usepackage{adjustbox}
\usepackage{soul}
\usepackage{caption}
\usepackage{subcaption}
\usepackage[normalem]{ulem}

\aclfinalcopy 

\title{AdvExpander: Generating Natural Language Adversarial Examples by Expanding Text}

\author{
    Zhihong Shao\textsuperscript{1}, Zitao Liu\textsuperscript{2}, Jiyong Zhang\textsuperscript{3}, Zhongqin Wu\textsuperscript{2}, Minlie Huang\textsuperscript{1}\thanks{*Corresponding author: Minlie Huang.}\\
    \textsuperscript{1} Department of Computer Science and Technology, Institute for Artificial Intelligence \\
    \textsuperscript{1} State Key Lab of Intelligent Technology and Systems \\
    \textsuperscript{1} Beijing National Research Center for Information Science and Technology \\
    \textsuperscript{1} Tsinghua University, Beijing 100084, China \\
    \textsuperscript{2} TAL Education Group, Beijing, China
    \textsuperscript{3} Hangzhou Dianzi University\\
    {\tt szh19@mails.tsinghua.edu.cn,  liuzitao@tal.com}\\
    {\tt jzhang@hdu.edu.cn, wuzhongqin@tal.com}\\ 
    {\tt aihuang@tsinghua.edu.cn}}

\date{}

\begin{document}
\maketitle
\begin{abstract}
Adversarial examples are vital to expose the vulnerability of machine learning models. Despite the success of the most popular substitution-based methods which substitutes some characters or words in the original examples, only substitution is insufficient to uncover all robustness issues of models. 
In this paper, we present \textbf{AdvExpander}, a method that crafts new adversarial examples by expanding text, 
which is complementary to previous substitution-based methods.
We first utilize linguistic rules to determine which constituents to expand and what types of modifiers to expand with. We then expand each constituent by inserting an adversarial modifier searched from a CVAE-based generative model which is pre-trained on a large scale corpus.
To search adversarial modifiers, we directly search adversarial latent codes in the latent space without tuning the pre-trained parameters.
To ensure that our adversarial examples are label-preserving for text matching, we also constrain the modifications with a heuristic rule.
Experiments on three classification tasks verify the effectiveness of AdvExpander and the validity of our adversarial examples. AdvExpander crafts a new type of adversarial examples by text expansion, thereby promising to reveal new robustness issues. 
\end{abstract}

\section{Introduction}
Adversarial examples are deliberately crafted from original examples to fool machine learning models, which can help (1) reveal systematic biases of data \cite{DBLP:conf/naacl/ZhangBH19,DBLP:journals/corr/abs-2004-02709}, (2) identify pathological inductive biases of models \cite{feng-etal-2018-pathologies} 
(e.g., adopting shallow heuristics \cite{DBLP:conf/acl/McCoyPL19} which are not robust and unlikely to generalize beyond training data), (3) regularize parameter learning \cite{DBLP:conf/conll/Minervini018}, (4) and evaluate stability \cite{DBLP:conf/acl/ChengJM19} 
or security level of models in practical use.

The most prevalent and effective practice of crafting natural language adversarial examples for classification is to flip characters \cite{DBLP:conf/acl/EbrahimiRLD18} or substitute words with their typos \cite{DBLP:conf/sp/GaoLSQ18,DBLP:conf/ijcai/0002LSBLS18}, synonyms \cite{DBLP:conf/milcom/PapernotMSH16,DBLP:conf/emnlp/AlzantotSEHSC18,DBLP:conf/acl/RenDHC19,DBLP:journals/corr/abs-1907-11932} or other context-compatible words \cite{DBLP:conf/acl/ZhangZML19}, 
while reusing the labels of the original examples as long as perturbations are few enough. However, these adversarial attacks limit the search space to the neighborhood of the original text and introduce only small lexical variation, which may not be able to uncover all robustness issues of models.

\begin{figure}[t]
    \includegraphics[width=0.48\textwidth]{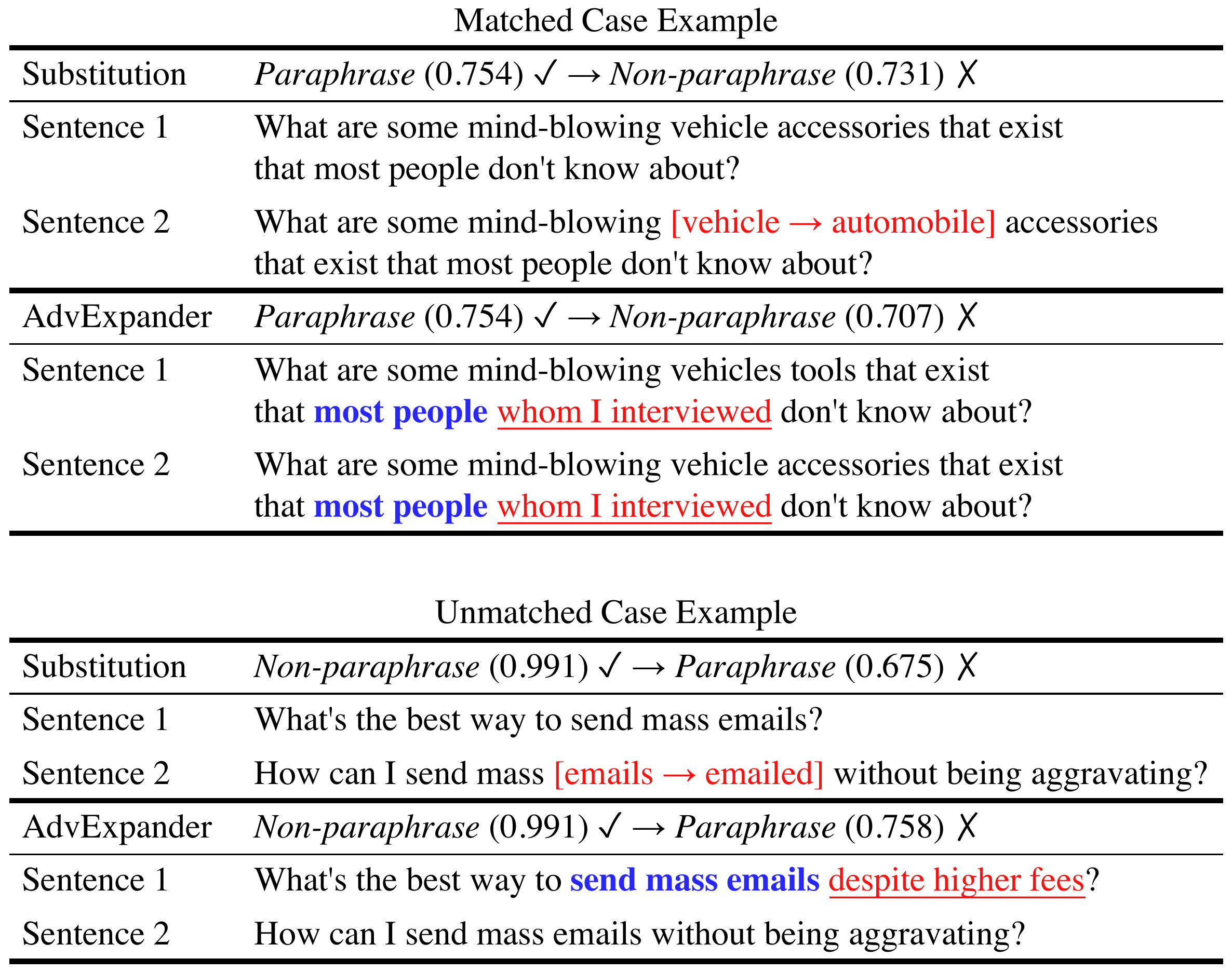}
    \caption{Adversarial samples on Quora Question Pairs, crafted against BERT by a substitution-based attack method and the insertion-based AdvExpander. 
    \textcolor{red}{$[A \rightarrow B]$} means substituting word \textit{A} with word \textit{B}. The underlined expressions are \textcolor{red}{\underline{adversarial modifiers}} inserted to expand the \textcolor{blue}{\textbf{target constituents}} in bold.}
    \label{fig:Quora_example}
\end{figure}

In this work, we present \textbf{AdvExpander}, which crafts new adversarial examples for classification by expanding text under the black-box setting. Specifically, we \textbf{first} use linguistic rules to identify constituents that are safe to expand without leading to an ill-formed text structure.
We \textbf{then} expand each constituent by inserting an adversarial modifier which is searched from a CVAE-based (Conditional Variational Auto-Encoder \cite{DBLP:conf/nips/SohnLY15}) generative model pre-trained on the Billion Word Benchmark \cite{DBLP:conf/interspeech/ChelbaMSGBKR14}. 
We search adversarial modifiers using REINFORCE \cite{DBLP:journals/ml/Williams92}. However, we avoid tuning pre-trained parameters as it can easily sacrifice grammaticality; instead, we additionally introduce a lightweight feed-forward network to search adversaries in the latent space.
To make AdvExpander applicable to text matching (e.g., natural language inference and paraphrase identification) besides text classification (e.g., sentiment classification), we design a heuristic rule to ensure modifications are label-preserving: for \textit{matched} cases (e.g., \textit{entailment} pairs in natural language inference and \textit{paraphrase} pairs in paraphrase identification), we only expand shared constituents in both texts with the same modifiers (see the \textit{matched} case in Fig \ref{fig:Quora_example}). 

We characterize AdvExpander in two aspects. 
\textbf{First}, AdvExpander differs from the aforementioned substitution-based attacks considerably. As the semantics of modifiers is far less restricted, and the expressions can be much more diverse than lexical substitutes, AdvExpander has larger search space and can introduce more linguistic variations besides lexical variation, e.g., syntactic variation and semantic variation. Take the \textit{matched} case in Fig \ref{fig:Quora_example} for example. For most existing substitution-based attack methods, the candidate substitutes of \textit{``vehicle''} are restricted to its synonyms, e.g., \textit{``automobile''}, \textit{``car''}. By contrast, for AdvExpander, there exist many reasonable modifiers of different types for \textit{``most people''}, e.g., clauses like \textit{``whom I interviewed''}, and prepositional phrases like \textit{``in the neighborhood''}. Therefore, AdvExpander is promising to measure the generalization ability of models. 
\textbf{Second}, as AdvExpander and substitution-based attacks adopt different types of manipulations (ours is based on insertion) and search adversarial examples in different search spaces, they complement each other and can be combined to boost attack performance. 

We applied AdvExpander to attack three state-of-the-art models (including RE2 \cite{DBLP:conf/acl/YangZGJC19}, BERT \cite{DBLP:conf/naacl/DevlinCLT19}, and WCNN \cite{DBLP:conf/emnlp/Kim14}) and two models with certified robustness to adversarial word substitutions \cite{DBLP:conf/emnlp/JiaRGL19} (including bag-of-words and CNN) on SNLI \cite{DBLP:conf/emnlp/BowmanAPM15}, Quora Question Pairs\footnote{https://data.quora.com/First-Quora-Dataset-Release-QuestionPairs}, and IMDB\footnote{https://datasets.imdbws.com/} which are commonly used datasets for natural language inference, paraphrase identification, and text classification respectively. We successfully reduce the accuracy of all target models to significantly below-chance level. Furthermore, the validity of our adversarial examples is verified by human evaluation. 

Our contributions are summarized as follows:
(1) We propose AdvExpander, which generates new adversarial examples by expanding constituents in texts with modifiers. This method is able to introduce rich linguistic variations and differs substantially from existing substitution-based methods; 
(2) On three classification datasets, AdvExpander substantially degrades the performance of three state-of-the-art models and two models robust to word substitutions, while human annotators remain highly accurate on such adversarial examples, which verifies the validity of our method.
    

\begin{figure*}[!htp]
    \centering
    \includegraphics[width=\textwidth]{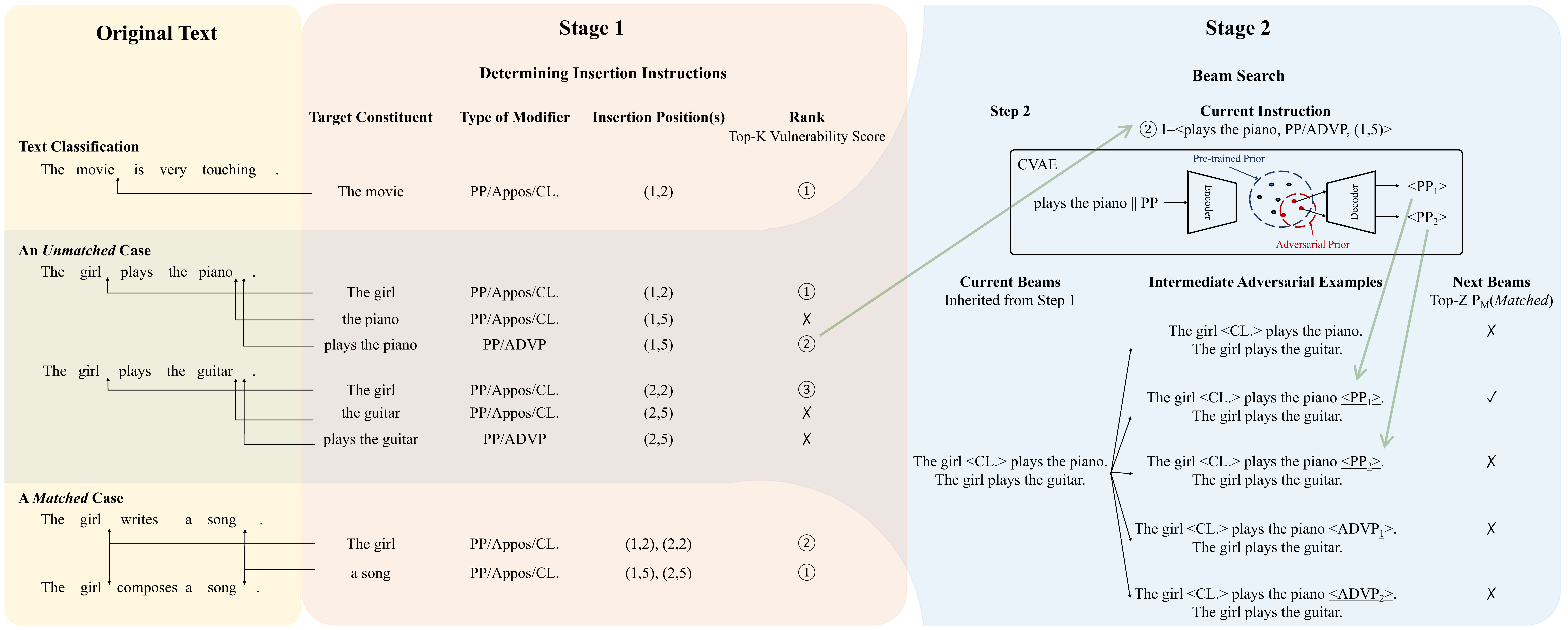}
    \caption{Workflow of AdvExpander for text classification and text matching.  
    \textit{``$\langle$CL.$\rangle$''}, \textit{``$\langle$PP$\rangle$''}, and \textit{``$\langle$ADVP$\rangle$''} denote a modifier of type \textit{CL.}, \textit{PP}, and \textit{ADVP}, respectively. \textbf{Stage 1} mainly determines the insertion instructions, i.e., which constituents to expand, what types of modifiers to expand with, and where to insert the modifiers (we use \textit{(i, j)} to indicate that the output modifier should be inserted after the $j^{th}$ word of the $i^{th}$ sentence). \textbf{Stage 2} is to search adversarial examples via beam search; in the beginning, beams are initialized as the original example. Each beam search step follows one instruction to search adversarial modifiers from a pre-trained CVAE. In this figure, beam size is 1; \textit{``$\langle \text{CL.} \rangle$''} is the modifier inserted at beam search step 1; \underline{underlined modifiers} (e.g., \textit{``\underline{$\langle \text{PP}_1 \rangle$}''}) are newly inserted modifiers at beam search step 2. If some successful adversarial example(s) is/are found after a beam search step, AdvExpander stops the beam search and returns the successful adversarial example with the lowest perplexity scored by GPT-2.}
    \label{fig:flowchart}
\end{figure*}

\section{Related Work}
Adversarial examples are of high value as they can reveal robustness issues of very successful deep classification models. According to how adversarial examples are crafted, recent work can be roughly divided into generation-based ones and edit-based ones.

\subsection{Generation-based Adversarial Examples}
Some studies utilize rules or neural generation methods to craft adversarial examples. \cite{DBLP:conf/acl/McCoyPL19} focused on natural language inference and generates an adversarial hypothesis from a premise based on linguistic rules. Though effective, rule-based methods introduce limited variations. \cite{DBLP:conf/naacl/IyyerWGZ18} introduced syntactic variation by paraphrasing original text with syntax-controlled network. The generated examples are not optimized to be adversarial.  \cite{DBLP:conf/acl/HovyKSK18} utilized Generative Adversarial Nets \cite{DBLP:conf/nips/GoodfellowPMXWOCB14} with generator generating adversarial examples and discriminator being the target model. This method is hard to balance grammaticality and adversary. \cite{DBLP:conf/iclr/ZhaoDS18} trained an inverter to map a text to a latent representation and searched adversaries nearby with heuristic rules. They trained the inverter on the original dataset, which might be insufficient to learn a smooth latent space. Also, their search strategy still has space for improvement. AdvExpander also involves neural generation. By contrast, we choose not to generate a complete text but generate only modifiers which are easier to control and thus less prone to syntactic errors. To learn smooth latent representations of texts, we pre-train a generative model on a large scale corpus. To balance grammaticality, efficiency and effectiveness when finding adversaries, we do not optimize pre-trained parameters but additionally introduce a lightweight feed-forward network to search adversaries in the latent space.

\subsection{Edit-based Adversarial Examples}
Most studies craft adversarial examples by editing the original text. 
Substitution is the most popular edit type. Substitution-based attacks are to search an optimal combination of adversarial substitutions under constraints. Under the black-box setting, adversarial attacks often involves scoring the importance of tokens (characters or words), which helps focus attention on important ones to reduce queries and perturbations. A common way of importance scoring is to measure changes of target model output after removing a token 
\cite{DBLP:journals/corr/abs-1805-12316}. The optimization process can be conducted by substituting tokens (1) in word order \cite{DBLP:conf/milcom/PapernotMSH16}
, (2) from important ones to less important ones 
\cite{DBLP:conf/acl/RenDHC19,DBLP:journals/corr/abs-1907-11932,DBLP:journals/corr/abs-2004-09984}, (3) with beam search \cite{DBLP:conf/acl/EbrahimiRLD18}, (4) or with population-based methods \cite{DBLP:conf/emnlp/AlzantotSEHSC18,DBLP:conf/acl/ZangQYLZLS20}. Constraints can be grammaticality, semantics-preservation, or context compatibility \cite{DBLP:conf/emnlp/bertattack}.

Compared with substitution which mainly introduces lexical variation, insertion and deletion together are likely to introduce richer variations but are far less popular, as they are more likely to render an adversarial example invalid. \cite{Wallace_2019} inserted universal triggers into text but the inserted strings are meaningless. \cite{DBLP:conf/acl/ZhangZML19} supported all three edit types but on token level, thus limited to small perturbations. Most relevantly, \cite{DBLP:conf/ijcai/0002LSBLS18} inserted adversarial phrases which were crafted manually. 
To the best of our knowledge, AdvExpander is the first efficient method that can automatically insert complex expressions, i.e., modifiers of constituents.

\section{Method}
\section{Task Definition}
Suppose a classifier $M$ maps the input text space $\mathcal{X}$ to the label space $\mathcal{Y}$. Let $X=s_1s_2...s_N$ be an input text and $Y$ is its label. For text classification, $X$ is a text with $N$ sentences and $s_i$ is the $i^{th}$ sentence. For text matching, $N=2$ and $<s_1, s_2>$ is the pair to be classified. Our goal is to craft a valid adversarial input $X_{adv}$ by expanding $X$, so that $M(X_{adv}) \neq Y$. Under the black-box setting, we only have access to the target model's predictions and the confidence scores.

\subsection{Overview}
Our method (Fig \ref{fig:flowchart}) can be divided into two stages. 
For convenience, we first introduce the concept ``insertion instruction'': an insertion instruction specifies one constituent to expand and the feasible type(s) of modifier to expand with.
At \textbf{the first stage}, we utilize linguistic rules to analyze the feasible insertion instructions, while ensuring that the insertions will not render the text structure ill-formed.
We consider a text structure ill-formed
if it is syntactically incorrect or there exists a constituent having multiple modifiers of the same constituency type (e.g., ``\textbf{The man} \underline{in white} \underline{behind the door} ...'').
To ensure that insertions are label-preserving for text matching, we further process the instructions so that we only expand shared constituents in both sentences with the same modifiers for \textit{matched} cases (e.g., \textit{entailment} pairs in natural language inference and \textit{paraphrase} pairs in paraphrase identification).
For computational efficiency, we only keep the most promising instructions (Eq. \ref{eq:potential}).

\textbf{The second stage} follows these instructions to find adversarial examples via beam search. 
At each beam search step, we follow one instruction to search adversarial modifiers in the latent space of a pre-trained generative model. We craft an adversarial example by inserting adversarial modifiers into the original example. In the end, we measure the perplexity of each successful adversarial example with GPT-2\footnote{https://s3.amazonaws.com/models.huggingface.co/bert/gpt2-pytorch\_model.bin} \cite{radford2019language}, and return the top-ranking one which is expected to be the most syntactically correct.

\subsection{Stage 1: Determining Insertion Instructions}
AdvExpander expands text by adding modifiers. 
We consider four types of modifiers, namely adverb phrase (ADVP), prepositional phrase (PP), appositive (Appos), and clause (CL.). For each input sentence $s_i$, we first obtain its constituency structure\footnote{https://s3-us-west-2.amazonaws.com/allennlp/models/elmo-constituency-parser-2018.03.14.tar.gz}, and then utilize handcrafted parsing templates\footnote{For more details, refer to supplementary material.} to determine which constituents to expand and what types of modifiers to expand with. 
To avoid rendering the text structure ill-formed, we ignore those types of modifiers that the target constituents already have.
These analytical results are formatted as a sequence of insertion instructions. Each instruction $I$ is defined as:
\begin{equation}
    \small
    \label{eq:instr}
    I=\langle c, t, P \rangle
\end{equation}
which means a modifier of type $t$ (one of the four types mentioned above) should be inserted into every position within the set $P$ to modify the target constituent $c$. 
For example, the \textit{unmatched} case in Fig \ref{fig:flowchart} has three feasible insertion instructions for each sentence. One of the instructions is \textit{I=$\langle$c=``The girl'', t=``PP/Appos./CL.'', P=\{(1,2)\}$\rangle$}, where \textit{(1,2)} means that the modifier should be inserted after the 2\textsuperscript{nd} word in the 1\textsuperscript{st} sentence.

To ensure insertions are label-preserving for text matching, we only expand shared constituents in both texts with the exact same modifiers for \textit{matched} cases. 
Take the \textit{matched} case in Fig \ref{fig:flowchart} for example. We only expand the shared constituents --- \textit{``The girl''} and \textit{``a song''} --- with the exact same modifiers, but ignore the different verb phrases \textit{``writes a song''} and \textit{``composes a song''}. Therefore, the instruction associated with \textit{``The girl''} has insertion position set \textit{P=\{(1,2),(2,2)\}}. 


For computational efficiency, we only retain those instructions with top-$K$ vulnerability score:

{\small
\begin{gather}
    \begin{aligned}
    \mathcal{I} = \text{Top-K}_I Score(I)
    \end{aligned}\\
    \begin{aligned}
    Score(I) = \max_{X_{adv} \in BS\_step(I, \{X\})} 1 - P_M(Y|X_{adv})
    \end{aligned}\label{eq:potential}
\end{gather}}
where $BS\_step(I, \{X\})$ (see the next section) returns $S$ adversarial examples searched in one-step beam search which follows the instruction $I$ and starts from $X$. $P_M(Y|X_{adv})$ is the probability of $Y$ given the intermediate adversarial example $X_{adv}$. For an instruction $I$, vulnerability score measures the vulnerability of the target constituent by insertion trials. We can calculate vulnerability score for each instruction in parallel.

\subsection{Stage 2: Searching Adversarial Examples}
\label{search}
The second stage follows the insertion instructions in decreasing order of vulnerability score to search adversarial examples via beam search.
During beam search, we maintain a set of intermediate adversarial examples $B$. 
At each beam search step, we follow one instruction $I$, and search adversarial modifiers for each $X_{adv} \in B$ from the latent space of a CVAE-based generative model.

\subsubsection{Design of Generative Model}
\begin{figure}[htp]
    \centering
    \includegraphics[width=0.25\textwidth]{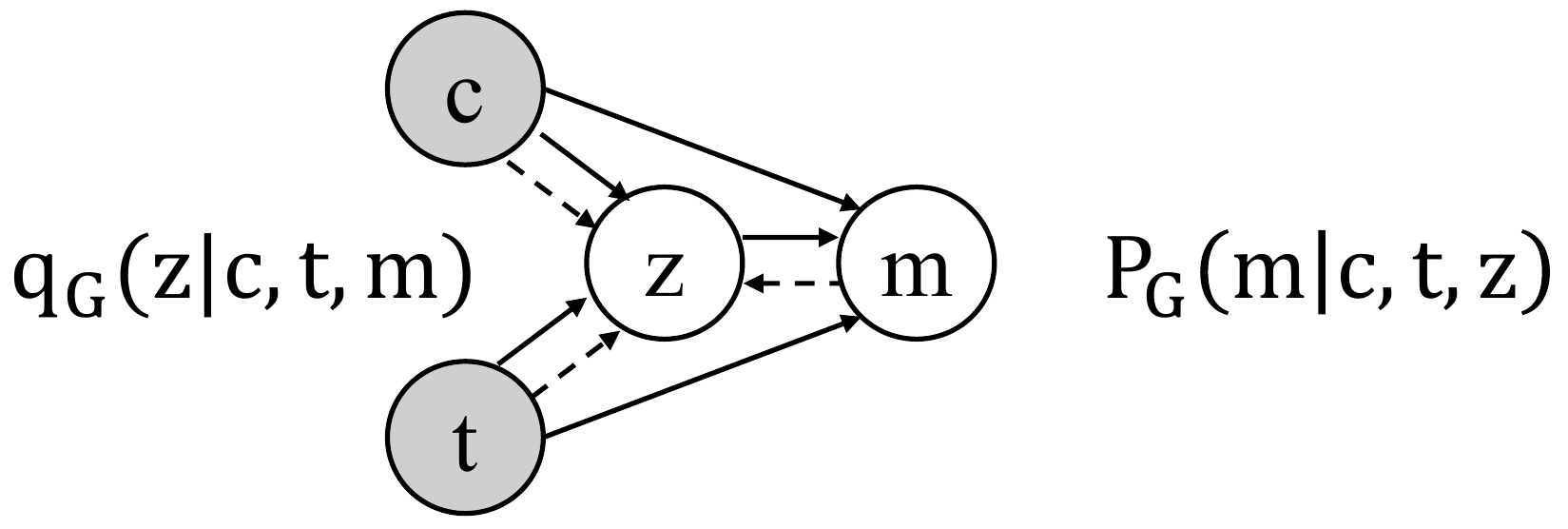}
    \caption{Graphical model of our CVAE-based generative model, where $c$, $t$, $z$, and $m$ denote target constituent, type of output modifier, latent variable, and output modifier, respectively. Dashed arrows are connections for the posterior distribution $q_G(z|c, t, m)$.}
    \label{fig:graph}
\end{figure}
Fig \ref{fig:graph} shows the graphical model of our generative model. Our generative model takes as input the target constituent $c$ and the expected type of the modifier $t$, samples the latent variable $z$, and generates the modifier $m$. We encode $c$ and decode $m$ with bi- and uni- directional RNNs, respectively.

Our generative model has two critical designs. \textbf{First}, we choose CVAE instead of Sequence-to-Sequence (Seq2Seq) model \cite{DBLP:journals/corr/BahdanauCB14}. This is because Seq2Seq trained with maximum likelihood estimation mainly captures low-level variations of expressions \cite{DBLP:conf/aaai/SerbanSLCPCB17,DBLP:conf/emnlp/ShaoHWXZ19}, thus failing to provide rich candidates of adversarial modifiers. \textbf{Second}, we generate modifiers conditioned on the target constituent instead of the entire text. 
This makes the distribution of conditions denser, which is beneficial to learn a smoother latent space and to improve the diversity and quality of text. This design also mitigates the gap between pre-trained corpus and attacked corpora, so that we can apply one pre-trained generative model to attack classifiers on different datasets.

\subsubsection{Pre-training Generative Model}
To learn a smooth latent space, we pre-trained our generative model on the Billion Word Benchmark. For training data, we treated each sentence in this corpus as having been expanded, and utilize the parsing templates used in stage 1 to extract constituents and their modifiers.

The loss function $\mathcal{L}^{pt}$ is the sum of two terms:
\begin{equation}
    \small
    \begin{split}
        \mathcal{L}^{pt} = \mathcal{L}_1^{pt} + \mathcal{L}_2^{pt}
    \end{split}
\end{equation}

The first term $\mathcal{L}_1^{pt}$ is the negative evidence lower bound of $\log P_G(m|c, t)$ which is log likelihood of modifier $m$ given its type $t$ and the modified constituent $c$:
\begin{equation}
    \small
    \begin{split}
        \mathcal{L}_1^{pt} = &- E_{q_G(z|c, t, m)}[\log P_G(m|c, t, z)] \\
        &+ D_{KL}(q_G(z|c, t, m)||p_G(z|c, t))
    \end{split}
\end{equation}
where $p_G(z|c, t)$ and $q_G(z|c, t, m)$ are the prior and posterior distribution of the latent variable $z$ respectively, which are isotropic Gaussian density function. $D_{KL}(\cdot)$ is KL divergence.

The second term $\mathcal{L}_2^{pt}$ is the loss of reconstructing $t$ from the latent variable $z$, which encodes $t$ into $z$ so that the type of output modifier can be better controlled \cite{DBLP:conf/acl/HuangKGx18}.
\begin{equation}
    \small
    \begin{split}
        \mathcal{L}_2^{pt} = - E_{q_G(z|c, t, m)}[\log P_G(t|z)]
    \end{split}
\end{equation}

\subsection{Beam Search}
Before beam search, the set of intermediate adversarial examples is initialized as $B_1 = \{X\}$. Let $I_i$ be the $i^{th}$ promising instruction in $\mathcal{I}$ according to its vulnerability score. The $i^{th}$ beam search step focuses on $I_i$ to expand $X_{adv} \in B_i$. The next beams $B_{i+1}$ is updated as follows:
\begin{equation}
    \small
    \begin{split}
        B_{i+1} = \text{Top-Z}_{X_{adv}^{'} \in B_i \cup BS\_step(I_i, B_i)} 1 - P_M(Y|X_{adv}^{'})
    \end{split}
\end{equation}
where $BS\_step(I_i, B_i)$ is a beam search step; following $I_i$, for each $X_{adv} \in B_i$, it searches $S$ modifiers from the generative model and separately inserts them into $X_{adv}$, resulting in $S$ new intermediate adversarial examples.

A straightforward way to search adversarial modifiers is to randomly sample $S$ latent codes from the prior distribution and decode a modifier $m_{adv}$ for each code:
\begin{equation}
    \small
    \begin{split}
        m_{adv} = arg\max_{m} P_G(m|c, t, z),\ z \sim P_G(z|c, t)
    \end{split}
\end{equation}

As the generative model is capable of producing a rich set of diverse modifiers, this method shows good attack performance. However, this searching process takes no consideration of the target model. We can further optimize the attack performance using REINFORCE\cite{DBLP:journals/ml/Williams92}.

To avoid sacrificing grammaticality for adversary, we choose not to finetune the pre-trained parameters of our generative model but directly search latent codes that will produce adversarial modifiers. As the prior network of our CVAE-based model maps an input to a space of proper but not necessarily adversarial latent codes, we additionally introduce an adversarial prior network, which is a trainable lightweight feed-forward network that narrows the latent space down to adversarial region (see Stage 2 in Fig \ref{fig:flowchart}).

The adversarial prior network computes adversarial prior  distribution $q_G^{adv}(z|c, t)$ which is isotropic Gaussian density function.
The adversarial prior network is initialized with the parameters of the pre-trained prior network, and is finetuned with REINFORCE. The reward is defined as:
\begin{equation}
    \small
    \begin{split}
        m_{adv} &= arg\max_m P_G(m|c, t, z),\ z \sim P_G^{adv}(z|c,t) \\
        R(z) &= - \log (P_M(Y|m_{adv} \rightarrow X_{adv}) + \alpha)
    \end{split}
\end{equation}
where $m_{adv}$ is the modifier decoded from the latent code $z$ which is sampled from the adversarial prior distribution, {\small $m_{adv} \rightarrow X_{adv}$} is the intermediate adversarial example crafted by inserting $m_{adv}$ into {\small $X_{adv}$} ({\small $\in B_i$}), $\alpha$ is a hyperparameter. To restrict that adversarial latent codes lie in the pre-trained prior distribution and produce modifiers of expected types, we regularize the finetuning procedure with:
\begin{equation}
    \small
    \begin{split}
        \Lambda = &D_{KL}(q_G^{adv}(z|c, t)||p_G(z|c, t)) \\
        &- E_{q_G^{adv}(z|c, t)}[\log{P_G(t|z)}]
    \end{split}
\end{equation}
Therefore, the total loss is:
\begin{equation}
    \small
    \begin{split}
        \mathcal{L}^{adv} = - E_{q_G^{adv}(z|c, t)} [R(z)] + \gamma \Lambda
    \end{split}
\end{equation}
The beam search step {\small $BS\_step(I_i, B_i)$}, for each {\small $X_{adv} \in B_i$}, minimizes {\small $\mathcal{L}^{adv}$} for $S$ steps, and returns $S$ new adversarial example (one example per step). 
Each {\small $X_{adv} \in B_i$} can be processed in parallel.

\subsubsection{Finalization of Adversarial Examples}
If some intermediate adversarial example {\small $X_{adv}^{'} \in BS\_step(I_i, B_i)$} fools the target model, we return the adversarial example with the lowest perplexity:
\begin{equation}
    \small
    \begin{split}
        X_{adv}^{*} = arg\min_{\substack{X_{adv}^{'} \in BS\_step(I_i, B_i)\\
        M(X_{adv}^{'}) \neq Y}} perplexity(X_{adv}^{'})
    \end{split}
\end{equation}
otherwise we continue beam search.

\section{Experiments}

\subsection{Tasks}
We evaluated AdvExpander on three datasets:
(1) \textbf{SNLI}:
A large scale dataset for natural language inference which is to judge whether a premise entails, contradicts or is independent of a hypothesis. The train/validation/test split is 550,152/10,000/10,000, respectively.
(2) \textbf{QQP}:
Quora Question Pairs for paraphrase identification which is to identify whether two sentences are paraphrases. The train/validation/test split is 384,348/10,000/10,000, respectively \cite{DBLP:conf/ijcai/WangHF17}.
(3) \textbf{IMDB}:
Movie reviews for document-level two-way sentiment classification, with 25,000/25,000 training/test instances respectively.

\subsection{Target Models}
We attacked RE2 and BERT on both SNLI and QQP, and attacked WCNN and BERT on IMDB. To verify that AdvExpander crafts new adversarial examples, we also attacked two models with certified robustness to adversarial word substitution, i.e., RBOW and RCNN from \cite{DBLP:conf/emnlp/JiaRGL19}.

(1) \textbf{RE2}:
A simple but effective model
which exploits rich alignment features for text matching. 
%
(2) \textbf{BERT}:
Bidirectional Transformer encoder which is pre-trained on large scale corpora. We finetuned BERT\textsubscript{base-uncased} 
on the three datasets respectively.
%
(3) \textbf{WCNN}:
Word-based Convolutional Neural Network.
%
(4) \textbf{RBOW}:
A robustly trained classifier with bag-of-words encoding. 
%
(5) \textbf{RCNN}:
A robustly trained bag-of-words model with input word vectors transformed by a CNN layer. 

\subsection{Substitution-based Attack Algorithms}
To verify that we can craft new adversarial examples, we compare AdvExpander with three recently proposed black-box substitution-based attack algorithms, i.e., PWWS, BERT-Attack, and TextFooler.
The three algorithms mainly differ in their estimation of word importance and the source of substitutes. 
As the three algorithms are demonstrated to be more effective than or comparable to many other algorithms, they are representative.

\noindent
\textbf{PWWS}:
\cite{DBLP:conf/acl/RenDHC19}
crafts semantic-preserving adversarial examples by replacing words with their synonyms (using WordNet\footnote{https://wordnet.princeton.edu/}) or replacing named entities with other similar ones. As PWWS has no named entity substitution rules specialized for SNLI or QQP, we applied PWWS on SNLI and QQP without named entity substitutions.

\noindent
\textbf{BERT-Attack}:
\cite{DBLP:conf/emnlp/bertattack} crafts adversarial examples by substituting (sub)words with context compatible alternatives sampled from BERT.

\noindent
\textbf{TextFooler}:
\cite{DBLP:journals/corr/abs-1907-11932} 
crafts semantic-preserving adversarial examples and finds synonyms with counter-fitting word embeddings \cite{mrksic-etal-2016-counter}.

\subsection{Implementation Details}
Insertion budget $K$ is 3/3/5 for SNLI/QQP/IMDB, respectively. Search steps $S$ is 80 and beam size $Z$ is 5. Thus, AdvExpander queries a target model for no more than 240/240/400 times to craft an adversarial example on SNLI/QQP/IMDB, respectively.
\textbf{For detailed implementation details, ablation analyses, case study, and error analysis, refer to supplementary material.}

\subsection{Automatic Evaluation}
\label{sec:auto}
\begingroup
\setlength{\tabcolsep}{3pt} 
\renewcommand{\arraystretch}{1} 
\begin{table}[htb]
    \centering
	\adjustbox{max width=.49\textwidth}{
    \begin{tabular}{c|ccc|cc|ccc}
        \hline
		Dataset & \multicolumn{3}{c|}{SNLI} & \multicolumn{2}{c|}{QQP} & \multicolumn{3}{c}{IMDB}\\
        \hline
        Model & RE2 & BERT & RBOW & RE2 & BERT & WCNN & BERT & RCNN\\
        \hline
	    Ori. Test & 86.9 & 90.7 & 79.4 & 88.6 & 91.3 & 90.0 & 92.0 & 79.3\\
	    Adv. & 23.8 & 27.2 & 21.4 & 33.0 & 44.2 & 10.9 & 16.4 & 6.6\\
	    \hline
	    Adv. Len. & 35.4 & 36.8 & 32.7 & 46.3 & 50.8 & 274.3 & 290.7 & 268.0\\
	    Ori. Len. & 23.8 & 23.8 & 23.7 & 24.1 & 23.8 & 245.0 & 257.1 & 238.0\\
	    Ori. Test (long) & 86.6 & 89.9 & 77.3 & 95.0 & 96.4 & 89.3 & 87.9 & 78.8\\
	    \hline
    \end{tabular}
    }
    \caption{Automatic evaluation performance, including model accuracy on the original test examples (\textit{``Ori. Test''}) and model accuracy on the corresponding adversarial examples crafted by AdvExpander (\textit{``Adv.''}). \textit{``Adv. Len.''} and \textit{``Ori. Len.''} denote the average length of \textbf{successful} adversarial examples and the \textbf{corresponding} original examples before expansion, respectively. \textit{``Ori. Test (long)''} denotes model accuracy on the original test examples that are longer than the average length of successful adversarial examples.}
    \label{tab:auto}
\end{table}

\begingroup
\setlength{\tabcolsep}{3pt} 
\renewcommand{\arraystretch}{1} 
\begin{table}[htb]
    \centering
    \adjustbox{max width=.49\textwidth}{
    \begin{tabular}{c|ccc|cc|ccc}
        \hline
		Dataset & \multicolumn{3}{c|}{SNLI} & \multicolumn{2}{c|}{QQP} & \multicolumn{3}{c}{IMDB}\\
        \hline
        Model & RE2 & BERT & RBOW & RE2 & BERT & WCNN & BERT & RCNN\\
        \hline
	    PS & 28.7 & 36.3 & 41.2 & 49.2 & 53.5 & 7.9 & 29.1 & 14.8\\
	    BA & 10.1 & 12.8 & 17.6 & 30.3 & 36.6 & 0.8 & 12.1 & 9.0 \\
	    TF & 2.9 & 4.4 & 8.9 & 32.1 & 38.1 & 0.4 & 14.3 & 10.0\\
	    \hline
	    \hline
	    PS + BA & 5.5* & 8.1* & 11.4* & 29.6* & 35.6* & 0.6* & 11.0* & 3.0*\\
	    PS + TF & 1.6* & 3.0* & 6.4* & 31.4* & 37.0* & 0.1 & 13.1* & 4.1*\\
	    BA + TF & 1.5* & 2.3* & 4.5* & 26.6* & 32.8* & \textbf{0.0} & 10.5* & 2.7 \\
	    \hline
	    Ours + PS & 3.6* & 5.5* & 10.1* & 21.0* & 29.2* & 2.9* & 12.9* & 2.6*\\
	    Ours + BA & 1.6* & 1.9* & 5.6* & \textbf{15.5} & \textbf{22.7} & 0.4* & \textbf{7.2} & 1.9 \\ 
	    Ours + TF & \textbf{0.3} & \textbf{0.4} & \textbf{1.3} & 16.4* & 23.8* & 0.2 & 8.6* & \textbf{1.8}\\
	    \hline
    \end{tabular}
    }
    \caption{Comparison between AdvExpander and substitution-based attack methods. \textit{``PS''}/\textit{``BA''}/\textit{``TF''} denotes PWWS/BERT-Attack/TextFooler, respectively. All numbers are model accuracy on the adversarial examples crafted by different attack methods. We also report model accuracy under the attacks of any two combined methods (e.g., \textit{``PS + BA''}): an attack is successful if at least one algorithm fools the target model. Accuracy that is significantly higher than the lowest accuracy (in bold) is marked with * for p-value $<$ 0.05 according to bootstrap resampling~\cite{DBLP:conf/emnlp/Koehn04}.}
    \label{tab:comparison}
\end{table}
\endgroup

We evaluated AdvExpander on the entire test sets for SNLI and QQP but on 1,000 random test samples for IMDB, as texts in IMDB are hundreds of words long and even the baseline PWWS is slow. We measured model accuracy on the original test examples and the corresponding adversarial examples
respectively (Table \ref{tab:auto}). 

AdvExpander is effective in crafting adversarial examples; it degrades the accuracy of all target models substantially. 
For example, the accuracy of BERT drops from above 90\% to below 28\% on both SNLI and IMDB. 

As AdvExpander crafts adversarial examples by expanding text, we further investigate the influence of text length on model accuracy. As shown in Table \ref{tab:auto}, though AdvExpander makes input texts longer, the target models remain high accuracy on the original test examples that are longer than the average length of adversarial examples, indicating that text length is not the factor why AdvExpander is successful to fool target models.

\noindent
\textbf{Comparison with Substitution-based Attacks}
We further investigate the relationship between AdvExpander and previous substitution-based attack algorithms (Table \ref{tab:comparison}).

AdvExpander degrades model accuracy more remarkably than PWWS in the most cases, but less remarkably than BERT-Attack and TextFooler.
Note that we choose not to modify an example if any insertion will render the text structure ill-formed.
When ignoring those examples we choose not to modify, the accuracy of BERT under our attacks is 6.2\% on SNLI and 37.1\% on QQP, respectively, which is competitive with the performance of BERT-Attack and TextFooler.

To verify that AdvExpander crafts new adversarial examples compared with substitution-based methods, we attacked RBOW and RCNN which have certified robustness to word substitutions.
Due to robust training, the two models are even harder to fool than some structurally more advanced models for substitution-based attacks.
However, as RBOW and RCNN have rather simple architecture and are only trained to be robust to word substitutions, they are unsurprisingly easier to fool than the other models for AdvExpander.
Take IMDB for example. RCNN is significantly more accurate (10.0\%) than WCNN (0.4\%) under TextFooler's attacks, but is much less accurate (6.6\%) than WCNN (10.9\%) under our attacks. 
Therefore, certified robustness to word substitutions may not indicate robustness to insertion-based adversarial examples.

We also combined AdvExpander with a substitution-based method to attack the target models (Table \ref{tab:comparison}). Specifically, an attack is considered successful if at least one of the two method fools the target model. The combinatorial attacks consistently boost attack performance. 
In the most cases, the highest performance boost is brought by combining AdvExpander with a substitution-based method but not by combining two substitution-based methods.
In other words, AdvExpander can craft adversarial examples in a way substitution-based methods is incapable of. 
Thus, AdvExpander is complementaryx to substitution-based methods and is promising to reveal new robustness issues.

\subsection{Human Evaluation}
\begingroup
\setlength{\tabcolsep}{3pt} 
\renewcommand{\arraystretch}{1} 
\begin{table}[htb]
    \centering
    \begin{subtable}{0.49\textwidth}
    \centering
    \adjustbox{max width=\textwidth}{
    \begin{tabular}{c|cc|cc}
        \hline
        Dataset & \multicolumn{2}{c|}{QQP} & \multicolumn{2}{c}{IMDB} \\
        \hline
		Metrics & Accuracy & Grammaticality & Accuracy & Naturalness \\
        \hline
	    Ori. Sampled & 85.0 & 2.65 & 88.0 & 2.69\\
	    \hline
	    TF & 71.5 & 2.45 & 84.0 & 2.57\\
	    Ours & 80.0** & 2.39 & 84.5 & 2.65*\\
	    \hline
    \end{tabular}
    }
    \label{tab:human}
    \end{subtable}
    \caption{
    Human evaluation of adversarial examples against BERT, in terms of human accuracy and grammaticality/naturalness.
    \textit{``TF''} denotes TextFooler.
    \textit{``Ori. Sampled''} shows evaluation on the corresponding original test samples.
    Bootstrap resampling~\cite{DBLP:conf/emnlp/Koehn04} is used as significance test between the two methods. ** marks significantly better performance for p-value < 0.01, and * for p-value $<$ 0.05.}
    \label{tab:validity}
\end{table}

To verify the validity of our adversarial examples, we conducted human evaluation (Table \ref{tab:validity}). 
We randomly sampled 200 adversarial examples against BERT on QQP and IMDB, respectively. These samples are mixed with the corresponding original test samples and the corresponding adversarial examples crafted by TextFooler; each example is presented to three workers on Amazon Mechanical Turk to annotate its label and whether it is grammatical/natural\footnote{As IMDB reviews are informal and contain grammatical errors, we measure naturalness on IMDB.
} (3-point Likert Scale). For each example, we aggregated human-predicted labels with majority vote, and computed human accuracy as the consistency between the aggregated labels and the gold labels. We also computed the average grammaticality/naturalness score.

Human accuracy on the original examples and our adversarial examples is close.
By contrast, on TextFooler's adversarial examples, human accuracy drops to 71.5\% on QQP, mostly due to imperfection of synonym candidates (e.g., substituting ``\textit{mechanical}'' in ``\textit{mechanical engineer}'' with ``\textit{mechanised}''). Therefore, our adversarial examples are label-preserving at an acceptable level.
Moreover, the grammaticality/naturalness score of our adversarial examples is close to that of the original samples, indicating that our adversarial examples are of good quality. 
Overall, these results demonstrate the validity of our adversarial examples.

\subsection{Adversarial Training}
We separately retrained RE2 on SNLI
augmented with 80K adversarial examples crafted on the training set by AdvExpander and TextFooler, and tested their robustness on the original test set (Table \ref{tab:advtrain}). 

For both TextFooler and AdvExpander, adversarial training helps improve a model's robustness to the attack method it is trained with, and slightly improves model accuracy on the original test set. Notably, as the original training set is large, training models with more adversarial examples can further improve models' robustness.

We also observed that adversarially training RE2 with TextFooler can hardly improve accuracy under AdvExpander's attacks (23.8\% $\rightarrow$ 24.1\%), and vice versa (2.9\% $\rightarrow$ 3.1\%). After augmenting the training set with both AdvExpander's and TextFooler's adversarial examples (80K for each), we improved model accuracy under AdvExpander's attacks (23.8\% $\rightarrow$ 30.8\%) and TextFooler's attacks (2.9\% $\rightarrow$ 7.4\%). This indicates that AdvExpander can generate new adversarial examples, and can reveal robustness issues that TextFooler fails to reveal. They complement each other.

\begingroup
\setlength{\tabcolsep}{3pt} 
\renewcommand{\arraystretch}{1} 
\begin{table}[htp]
    \centering
    \adjustbox{max width=0.48\textwidth}{
    \begin{tabular}{c|c|c|c|c}
        \hline
		Training Set & Ori. Train & + TF & + Ours & + Ours \& TF \\
        \hline
	    Ori. Test & 86.9 & 87.3 (+0.4) & 87.0 (+0.1) & 87.2 (+0.3)\\
	    \hline
	    TF & 2.9 & 8.0 (+5.1) & 3.1 (+0.2) & 7.4 (+4.5)\\
	    Ours & 23.8 & 24.1 (+0.3) & 30.0 (+6.2) & 30.8 (+7.0)\\
	    \hline
    \end{tabular}
    }
    \caption{Model accuracy on the original test examples (\textit{``Ori. Test''}) and adversarial examples (\textit{``TF''} and \textit{``Ours''}) after adversarially training RE2 on SNLI with TextFooler (\textit{``+TF''}), with AdvExpander (\textit{``+Ours''}), or with both AdvExpander and TextFooler (\textit{``+Ours \& TF''}). 
    \textit{``TF''} stands for TextFooler.
    Numbers in parentheses are improvements over the model trained on the original training set (\textit{``Ori. Train''}).}
    \label{tab:advtrain}
\end{table}
\endgroup





\section{Conclusion}
In this paper, we present AdvExpander which generates new natural language adversarial examples by expanding text. Extensive experiments demonstrate the effectiveness of our algorithm and the validity of our adversarial examples. Our adversarial examples are substantially different from previous substitution-based adversarial examples, thus promising to reveal new robustness issues.

\bibliography{acl2020}
\bibliographystyle{acl_natbib}

\newpage

\end{document}